\documentclass[times,twocolumn,final]{elsarticle}

\usepackage{lineno,hyperref}
\usepackage{amsmath,amsfonts,amssymb}
\usepackage{graphicx}
\usepackage{multirow}
\usepackage{subcaption}
\usepackage{pdflscape}
\usepackage{setspace}
\usepackage{tocloft}
\usepackage[export]{adjustbox}

\biboptions{authoryear}

\begin{document}

\begin{frontmatter}

\title{OSCARS: An Outlier-Sensitive Content-Based Radiography Retrieval System}


\author[emorycs]{Xiaoyuan Guo}
\ead{xiaoyuan.guo@emory.edu}
\author[usc]{Jiali Duan}
\ead{jialidua@usc.edu}
\author[iupui]{Saptarshi Purkayastha}
\ead{saptpurk@iupui.edu}
\author[emorymed,emoryrad]{Hari Trivedi}
\ead{hari.trivedi@emory.edu}
\author[emorymed,emoryrad]{Judy Wawira Gichoya}
\ead{judywawira@emory.edu}
\author[mayo,asu]{Imon Banerjee\corref{mycorrespondingauthor}}
\cortext[mycorrespondingauthor]{Corresponding author}
\ead{banerjee.imon@mayo.edu}

\address[emorycs]{Department of Computer Science, Emory University, GA, USA}
\address[usc]{Ming Hsieh Department of Electrical and Computer Engineering,  University of Southern California, CA, USA}
\address[emorymed]{School of Medicine, Emory University, GA, USA}
\address[emoryrad]{Department of Radiology and Imaging Sciences, Emory University, GA, USA}
\address[iupui]{Indiana University-Purdue University Indianapolis, School of Informatics and Computing, IN, USA}
\address[mayo]{Department of Radiology, Mayo clinic, Phoenix, AZ, USA}
\address[asu]{School of Computing and Augmented Intelligence, Arizona State University, AZ, USA}

\begin{abstract}
Improving the retrieval relevance on noisy datasets is an emerging need for the curation of a large-scale clean dataset in the medical domain. 
While existing methods can be applied for class-wise retrieval (aka. inter-class), they cannot distinguish the granularity of likeness within the same class (aka. intra-class). The problem is exacerbated on medical external datasets, where noisy samples of the same class are treated equally during training. Our goal is to identify both intra/inter-class similarities for fine-grained retrieval.  
To achieve this, we propose an \textbf{O}utlier-\textbf{S}ensitive \textbf{C}ontent-based r\textbf{A}diologhy \textbf{R}etrieval \textbf{S}ystem (\textbf{OSCARS}),  consisting of two steps. First, we train an outlier detector on a clean internal dataset in an unsupervised manner. Then we use the trained detector to generate the anomaly scores on the external dataset, whose distribution will be used to bin intra-class variations. Second, we propose a quadruplet ($a$, $p$, $n_{intra}$, $n_{inter}$) sampling strategy, where intra-class negatives $n_{intra}$ are sampled from bins of the same class other than the bin anchor $a$ belongs to, while $n_{iner}$ are randomly sampled from inter-classes. We suggest a weighted metric learning objective to balance the intra and inter-class feature learning. We experimented on two representative public radiography datasets. Experiments show the effectiveness of our approach. The training and evaluation code can be found in \url{https://github.com/XiaoyuanGuo/oscars}.
\end{abstract}


\end{frontmatter}


\section{Introduction}
\label{sect:intro}  
With the widespread adoption of radiology in diagnosis and treatment planning, the amount of medical image data is rapidly increasing~\cite{hwang2012medical}. Fast and effective retrieval in large-scale medical image repositories has been demanding to support data management, research and clinical applications~\cite{sotomayor2021content}. One common way to retrieval images is content-based, which has been widely researched and applied to the medical field~\cite{wang2014learning,dubey2021decade,chowdhury2016efficient,chen2022deep}. 
\begin{figure}[t]
  \centering
  \includegraphics[width=\linewidth]{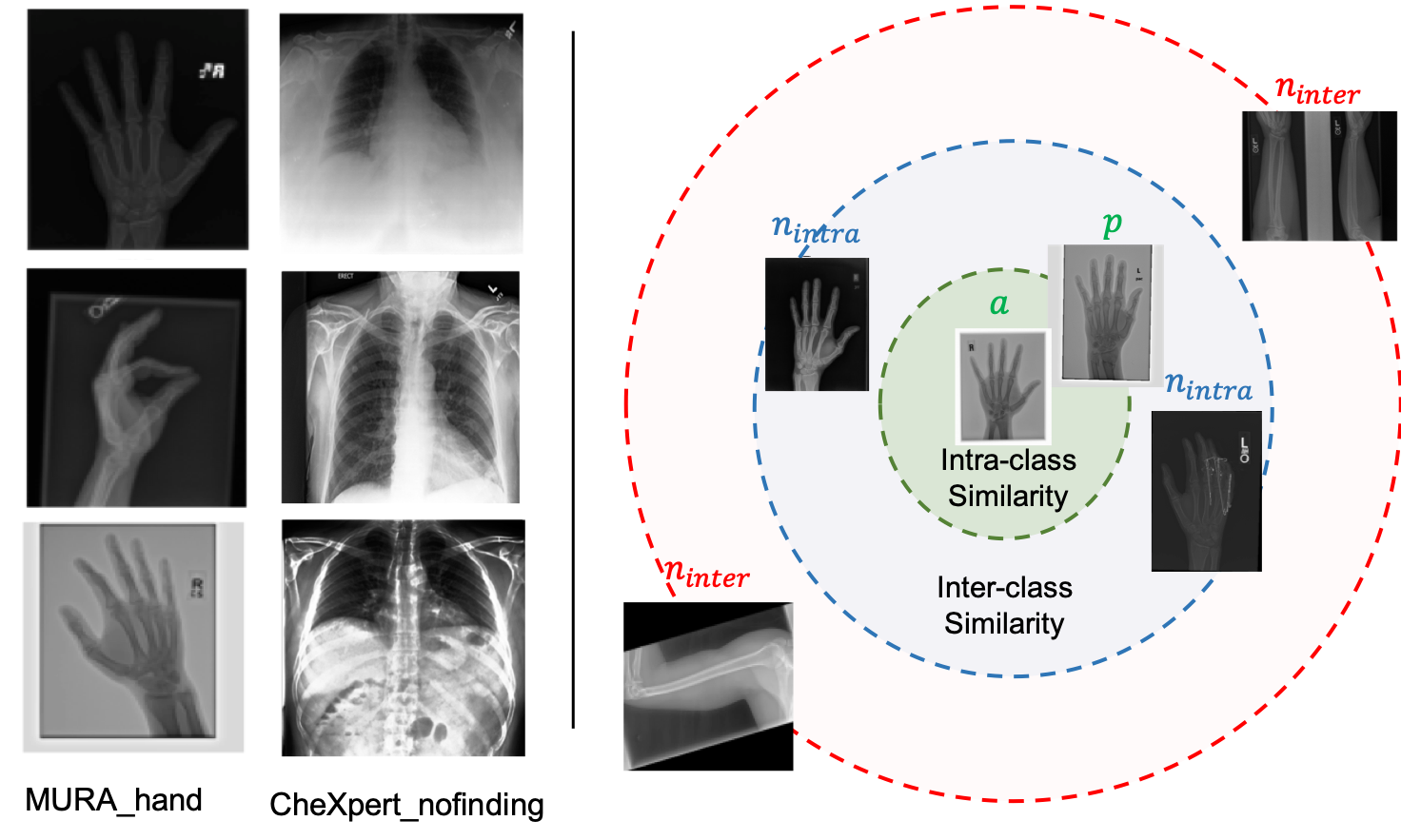}
  \caption{(Left) Examples of intra-class variations. 1st column shows samples from Stanford MURA \textit {HAND} class and 2nd column presents CheXpert\textit{No Finding} class data. (Right) Oscars learns the intra-class and inter-class similarity simultaneously. Images with intra-class similarity $p$ should be closer to the given image $a$ than the samples that show inter-class similarity $n_{intra}$ in the feature space.} \label{fig2}
\end{figure}

For a given query image, a content-based image retrieval (CBIR) system returns a ranked list of images from the database based on a similarity measure between the query and retrieved images~\cite{duan2021bridging,revaud2019learning}. 
The core idea behind CBIR is to minimize the distance of an anchor image $a$ to its positive counterparts $p$s and maximize the distance to the corresponding negative images $n$s in the feature space. Usually, the positive images are in the same class as the anchor image. However, adopting this strategy can be problematic as it only considers the inter-class variation. The assumption - as long as $a$ and $p$ are from the same class, they show similar visual features - is not realistic as samples from one class often exhibit certain intra-class variations. Noisy, under-represented data can exist, also called \emph {outliers}. This phenomenon is more common in radiology as images are often acquired via different equipment from different sources and varies based on acquisition protocols. These variations, as shown in the left part of Fig.~\ref{fig2}, pose specific challenges in the consumer domain and need to be recognized in assessing image similarity~\cite{akgul2011content}.

Although there have been multiple studies for radiograph retrieval, few of them pay attention to the intra-class similarity problem. 
~\cite{anavi2015comparative} investigated X-ray image retrieval with both distance-based and probability-based approaches. 
~\cite{chowdhury2016efficient} proposed a content-based medical image retrieval (CBMIR) system for radiographic images and employed a CNN to obtain high-level image representations. 
~\cite{qayyum2017medical} proposed a CBIR framework by training a CNN for the classification task. 
~\cite{layode2020chest} developed a chest X-ray image retrieval system for COVID-19 detection with deep denoising autoencoders as feature extractors. 
~\cite{zhong2021deep} designed an image retrieval system for COVID-19 chest radiograph via optimizing a multi-similarity loss. Outside medical domain, methods including FastAP~\cite{cakir2019deep}, MultiSimilarity~\cite{wang2019multi}, CircleLoss\cite{sun2020circle} and SupCon~\cite{khosla2020supervised} try to discover challenging negative data to improve the retrieval accuracy. Nevertheless, these existing efforts all emphasize the inter-class similarity but neglect the intra-class similarity.

In this paper, we focus on relevant radiograph image retrieval in external datasets which can contain lots of noisy data compared to the clean internal dataset. 
Such a system will help to \emph{collect cleaner external image dataset with minimal human effort and accelerate AI evaluation}. 
To achieve the goal, we propose an \textbf{O}utlier-\textbf{S}ensitive \textbf{C}ontent-based r\textbf{A}diologhy \textbf{R}etrieval \textbf{S}ystem (\textbf{OSCARS}), which takes both the intra-class and inter-class variations into consideration. To acquire the intra-class variation information, we adopt the unsupervised anomaly detectors trained on the internal dataset and utilize the assigned anomaly scores to the external dataset to split each class into several bins, with each bin in a certain range regarding the anomaly scores. Based on which, we construct the quadruplet data $(a, p, n_{intra}, n_{inter})$ with an anchor image $a$, a positive image $p$ from the same class and same bin, an intra-class negative image $n_{intra}$ from the same class but different bins, and an inter-class negative image $n_{inter}$ that is from a different class.

\begin{figure*}[tp]
  \centering
  \includegraphics[width=0.8\linewidth]{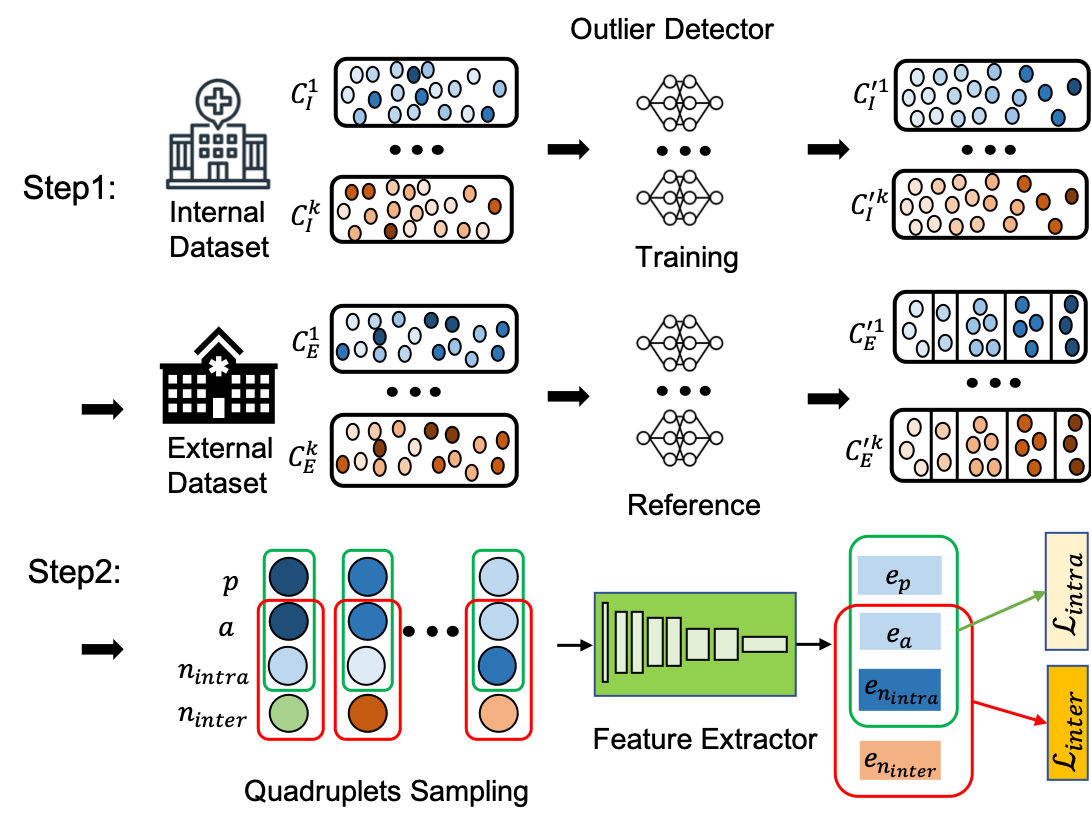}
  \caption{OSCARS architecture involves two main steps. Step1: train anomaly detectors on the internal dataset for each class $C^{i}_{I}$; learn clean in-distributions with anomaly scores assigned to $C^{'i}_{I}$; apply the trained anomaly detectors on each class $C^{i}_{E}$ of the external dataset and split the data into several bins $C^{'i}_{E}$ according to the anomaly scores. (Dark colors mean more distribution shifts.) Step2: generate quadruplets $(a, p, n_{intra}, n_{inter})$ by sampling the intra-class positive, negative and inter-class negative simultaneously; learn the intra-class and inter-class similarity in feature space with the intra-class triplet loss $L_{intra}$ and inter-class triplet loss $L_{inter}$.~\label{oscars}}
\end{figure*}

With the proposed quadruplet sampling strategy, we incorporate the intra-class discriminative information into the training data and hence improve the retrieval of sensitivity outlier-related queries after model training. All the images in a quadruplet are fed into the feature extractor to learn their latent embeddings $(e_{a}, e_{p}, e_{n_{intra}}, e_{n_{inter}})$. As illustrated in the right of Fig.~\ref{fig2}, we then learn the intra-class embedding similarity to achieve $(Sim(e_{a}, e_{p})$ $>$ $Sim(e_{a},$ $e_{n_{intra}}))$ with an intra-class triplet loss $L_{intra}$ and the inter-class similarity for $(Sim(e_{a}, e_{n_{intra}}) > Sim(e_{a}, e_{n_{inter}}))$ with an inter-class triplet loss $L_{inter}$ in a weighted way. Our summarized contributions are:
\begin{enumerate}
    \item We introduce the task of outlier-sensitive image retrieval for noisy external medical image dataset and propose an effective image retrieval system \textbf{OSCARS} to enhance the relevance of outlier-related results.
    \item We propose to acquire intra-class information of external datasets via anomaly detectors trained unsupervised. By training on clean internal datasets, the anomaly detectors assign each sample of the external dataset with a specific anomaly score. Based on which, we split each class into several bins with different intra-class variations.
    \item We sample both the intra-class and inter-class negative images to construct quadruplets for intra-class and inter-class similarity learning. 
    \item We demonstrate the model effectiveness with two public representative radiography datasets - Stanford Muscoloskeletal Radiography (MURA)~\cite{rajpurkar2017mura} and CheXpert~\cite{irvin2019chexpert}.  
\end{enumerate}

\section{Methodology}
Given a clean internal dataset $D_{I}$ and a noisy external dataset $D_{E}$, the external data of class $c$ can contain outliers visually different from the internal class. Therefore, a conventional image retrieval system for the external dataset will be insufficient as it merely treats all the samples from one class as the same without considering the intra-class variations. 
Thus, the system will lack sensitivity to the outliers, undermining the retrieval accuracy. 
Our objective is to train an image retrieval model that will prioritize the images with both intra-class and inter-class dissimilarity during retrieval ranking.  Figure~\ref{oscars} summarizes the whole framework of our model. There are mainly two steps involved. First, we design to learn intra-class information in an unsupervised way (introduced in Sec.~\ref{intra_bin}). Second, we propose to sample training data that are with intra-class bin information and inter-class information (introduced in Sec.~\ref{balance}). With these steps, images with the same labels and similar contents are pulled together by maintaining intra-class similarity.

\subsection{Learning intra-class information}~\label{intra_bin}
Due to the difficulties of collecting annotated data with intra-class information provided in the medical domain, the outlier-sensitivity research on medical images has been delayed. 
To overcome the problem, we propose to generate intra-class labels automatically inspired by a recent work -  MedShift~\cite{guo2021medshift}. Given a clean internal dataset $D_{I}$,  MedShift has suggested an approach to identify outliers for noisy external dataset $D_{E}$. Following the same steps of MedShift, we first obtain the internal distribution information by training an unsupervised outlier detector named CVAD~\cite{guo2021cvad} for each class on the same internal datasets used in~\cite{guo2021medshift}. Then, the trained anomaly detectors are evaluated on the external datasets as they have learnt intra-class discriminative features. Thus, each external data has its anomaly score, based on which we split each class into $B$ bins with the K-Means clustering techniques~\cite{lloyd1982least,macqueen1967some}. $B$ (5 in our paper) is determined by the Elbow method~\cite{thorndike1953belongs}. The resulting bins are in different anomaly score ranges. With the data from different bins, we get the intra-class labels. Given that both the intra-class and inter-class labels are available, for each image $a$, we randomly sample one intra-class positive image $p$, one intra-class negative sample $n_{intra}$ and one inter-class negative sample $n_{inter}$ accordingly, thus collecting the quadruplets $(a, p, n_{intra}, n_{inter})$ for training.

\subsection{Balancing the inter- and intra-class influence}~\label{balance}
With the sampled quadruplets data, we feed each of the image to a CNN-based feature extractor to acquire latent embeddings $(e_{a}, e_{p}, e_{n_{intra}}, e_{n_{inter}})$. For simplicity, we adopt the ResNet18~\cite{he2016deep} pre-trained on ImageNet~\cite{deng2009imagenet} as the network backbone. OSCARS is designed to consider both the inter-class similarity and the intra-class similarity at the same time, which brings the model advantages of acquiring the sensitivity of intra-class outlier relevance during image retrieval. However, balancing the effect of the two parts is a challenging problem. Too much weight on intra-class information will distract the general retrieval accuracy of inter-class data. Therefore, we design an intra-class triplet margin loss and an inter-class triplet margin loss to optimize the model architecture. To balance the influence of intra-class and intra-class information on final ranking, we adopt a weighted loss formulated as:
\begin{equation}~\label{loss}
\begin{split}
\mathcal{L} &= \lambda \mathcal{L}_{intra}(e_{a}, e_{p}, e_{n_{intra}}) + (1-\lambda) \mathcal{L}_{inter}(e_{a}, e_{n_{intra}}, e_{n_{inter}}) \\
&= \lambda(max\{d(e_{a}, e_{p}) - d(e_{a}, e_{n_{intra}}) + \mathcal{M}_{intra}, 0\}) \\
&+(1-\lambda) (max\{d(e_{a}, e_{n_{intra}}) - d(e_{a}, e_{n_{inter}}) + \mathcal{M}_{inter}, 0\})
\end{split}
\end{equation}
where $d(x_i, y_i)={\left \| x_{i}-y_{i} \right \|}_2$. $\lambda$, $\mathcal{M}_{intra}$ and $\mathcal{M}_{inter}$ are set as 0.05, 1 and 2 in our experiments respectively.

When we have a query image unseen during training, we first acquire the query representation with the trained image feature backbone and then compute the cosine similarity between the representative features of the query image and dataset images. Images are ranked based on the similarity scores in the descending order.

\section{Experiments}
We have evaluated our approach on two publicly available large-scale radiograph image datasets. The first is Stanford MURA dataset, a large dataset of bone X-rays, which contains seven classes - \textit{HAND}, \textit{FORARM}, \textit{FIGER}, \textit{SHOULDER}, \textit{ELBOW}, \textit{WRIST}, \textit{HUMERUS}. The second is CheXpert dataset, which in total has 14 classes - \textit{No Finding}, \textit{Enlarged Cardiomediastinum}, \textit{Cardiomegaly}, \textit{Lung Lesion}, \textit{Lung Opacity}, \textit{Edema}, \textit{Consolidation}, \textit{Pneumonia}, \textit{Atelectasis}, \textit{Pneumothorax}, \textit{Pleural Effusion}, \textit{Pleural Other}, \textit{Fracture}, \textit{Support Devices}. As the chest x-ray images are with two views - frontal and lateral. We here only use frontal view and leave the lateral view for future studies. See more details in the supplementary materials.

\begin{table*}[htp]
\caption{Quantitative performance on Stanford MURA and CheXpert datasets~\label{allinone}. Bold indicates the best.\vspace{-3mm}}
\resizebox{\linewidth}{!}{%
\begin{tabular}{|l|ccccc|ccccc|ccccc|ccccc|}
\hline
\multirow{2}{*} {Method}& \multicolumn{10}{c|}{MURA}  & \multicolumn{10}{c|}{CheXpert}\\ \cline{2-21} 
  & R@1$\uparrow$ & R@5$\uparrow$ & R@10$\uparrow$ & R@50$\uparrow$  & R@100$\uparrow$ & P@1$\uparrow$ & P@5$\uparrow$ & P@10$\uparrow$ & P@50$\uparrow$ & P@100$\uparrow$  & R@1$\uparrow$ & R@5$\uparrow$ & R@10$\uparrow$ & R@50$\uparrow$ & R@100$\uparrow$  & P@1$\uparrow$ & P@5$\uparrow$ & P@10$\uparrow$ & P@50$\uparrow$ & P@100$\uparrow$ \\ \hline 
DeepRank~\cite{wang2014learning} & 0.912 & 0.914  & 0.912  & 0.906 & 0.903 & 0.912  & 0.964  & 0.972  & 0.984  & 0.988  & 0.734 & 0.694 & 0.716 & 0.721  & 0.442 & 0.734 & 0.911 & 0.961 & 1.000 & 1.000\\
FastAP~\cite{cakir2019deep} & 0.927 & 0.930 & 0.931 & 0.932  & 0.933  & 0.927  & 0.956  & 0.961  & 0.973  & 0.977 & 0.734 & 0.742 & 0.733 & 0.716 & 0.710 & 0.734 & 0.943 & 0.968 & 1.000 & 1.000 \\
MultiSimilarity~\cite{wang2019multi} & 0.923  & 0.921 & 0.919  & 0.915 & 0.913 & 0.923  & 0.955 & 0.968  & 0.977 & 0.980  & 0.695 & 0.680 & 0.677 & 0.676  & 0.682  & 0.695 & \textbf{0.957} & \textbf{0.975} & 1.000 & 1.000 \\
CircleLoss\cite{sun2020circle} & 0.929 & 0.932 & 0.933 & 0.934 & 0.934 & 0.929  & 0.960 & 0.964  & 0.979  & 0.985 & 0.727 & 0.703 & 0.718 & 0.717 & 0.726  & 0.727 & 0.936 & 0.968 & 1.000 & 1.000  \\ 
SupCon~\cite{khosla2020supervised} & 0.930 & \textbf{0.933} & \textbf{0.936} & \textbf{0.938} & \textbf{0.937} & 0.930 & 0.964 & 0.971  & 0.981  & 0.985 & 0.776 & 0.730 & 0.720 & 0.734 & 0.726   & 0.776 & 0.936 & 0.950 & 1.000 & 1.000 \\ \hline
OSCARS \textit{(ours)} & \textbf{0.931} & 0.922 & 0.920  & 0.913  & 0.910 & \textbf{0.931}  & \textbf{0.965}  & \textbf{0.974}  & \textbf{0.986} & \textbf{0.991} & \textbf{0.787} & \textbf{0.763} & \textbf{0.747} & \textbf{0.745} & \textbf{0.743} & \textbf{0.787} & 0.908 & 0.950 & 1.000 & 1.000 \\   \hline
\end{tabular}%
}
\end{table*}

\subsection{Evaluation Metrics}
For the retrieval task, we report the retrieval \textit{recall} at rank $K$ ($R@K$, $K \in \{1, 5, 10, 50, 100\}$), \textit{precision} at rank $K$ ($P@K$, $K \in \{1, 5, 10, 50, 100\}$), outlier sensitivity ($S@K$, $K \in \{1, 5, 10, 50, 100\}$). 
The metric \textit{recall} is the percentage of relevant images retrieved over the total number of retrieved images, defined as $recall = \frac {N_{R}} {K}$
where $R$ represents the relevant images retrieved.
The metric \textit{precision} is assigned based on the existence of the same labels between the query image and the retrieved images. If $\delta(\cdot)\in\{0,1\}$ is an indicator function, the \textit{precision} is defined as $precision = \frac{\sum_{i=1}^{K} \delta({R}^{i}>0)}{K}$.
Additionally, we evaluate the outlier sensitivity by calculating the anomaly score difference with $sensitivity = \sum_{i=1}^{N_{R}}{\frac{|\mathcal{A}^{i}_{R}-\mathcal{A}_{q}|}{N_{R}}}$,
where $\mathcal{A}$ means anomaly score. We scale the anomaly scores of MURA dataset into [0,1] with the sigmoid function due to the large variations of its anomaly scores. 

\subsection{Implementation Details}
The pipelines are developed using Pytorch 1.9.0, Python 3.7.3 and Cuda compilation tools V11.4 on a machine with 4 NVIDIA Quadro RTX A6000 GPUs with 48GB memory. The training for all the models is run for 50 epochs with a start learning rate 0.001 and a SGD optimizer. 
\begin{figure*}[ht]
  \centering
  \includegraphics[width=\linewidth]{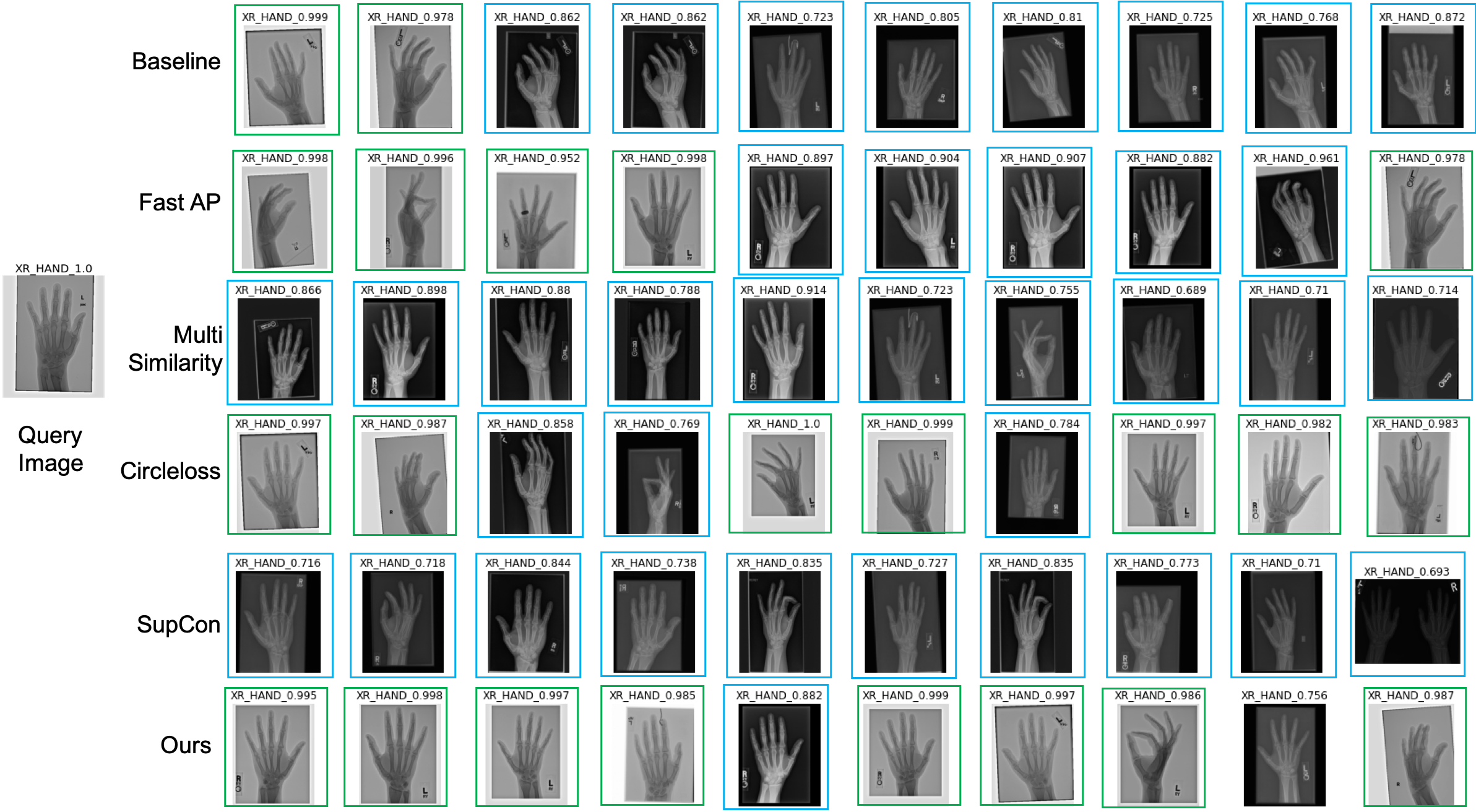}
  \caption{Hand results, left is the query image, right shows retrieval results. Green boxes mean both intra- and inter-class correct; blue boxes are for inter-class correct predictions. Each retrieval image has its label on top of itself. For correct predictions, we also put the anomaly scores on them. Closer anomaly scores mean more similarity. }
  \label{mura_res} 
\end{figure*}

\begin{figure*}[htp]
  \centering
  \includegraphics[width=\linewidth]{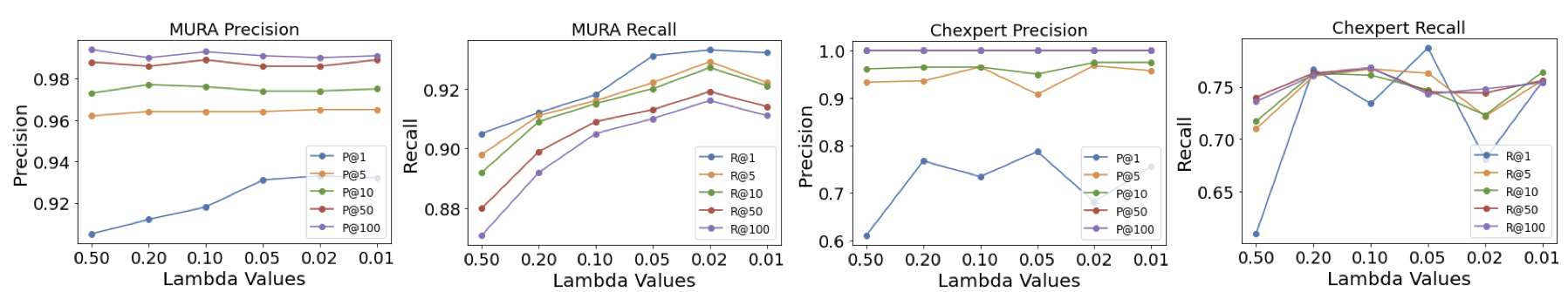}
  \caption{Effects with different lambda values on different datasets.}
  \label{lambda}
\end{figure*}

\subsection{Search Results}
As a representative image retrieval method with triplet data in training, we select DeepRank as our baseline. State-of-the-art CBIR approaches including FastAP, MultiSimilarity, CircleLoss and SupCon are used to compare the model performance. Notably, we keep the feature extractor consistent for all the methods to ensure fair comparisons.

\paragraph{Quantitative Results:}
Table~\ref{allinone} presents the recall and precision performance for both Stanford MURA and CheXpert datasets respectively. Since the data in CheXpert can have multiple labels, we calculate the correct hit with the strategy - loose match, which means that for a query chest X-ray with multiple labels, a retrieval image is relevant as long as it has one label matched. Compared to the baseline DeepRank, Oscars can enhance the recall and precision performances on both datasets and achieve the best recall at 1 and precision at 1. In general, SupCon has the highest recall for MURA dataset. Nonetheless, Oscars achieves the best precision for MURA dataset and recall for CheXpert. Additionally, we report the sensitivity results in the supplementary materials.

\paragraph{Qualitative Results:}
Figure ~\ref{mura_res} shows an example of a \textit{HAND} query image in MURA dataset. The corresponding retrieval results including ours are present in different rows. As can be seen, although many methods can achieve high recall and precision (see Table~\ref{allinone}), they fail to distinguish the intra-class variations. Especially, MultiSimilarity and SupCon exhibit little sensitivity to the noisy query. Comparatively, our method can prioritize intra-class similarity and rank the images with similar anomaly semantics ahead. 
Please refer to the supplementary materials for more results.

\paragraph{Impact of Lambda:} We also explore the impacts of applying different $\lambda$ values to the loss function (Eqn.~\ref{loss}). A good balance between the intra-class and inter-class information will enable the retrieval system to acquire both accurate inter-class and outlier-sensitive intra-class results. Figure~\ref{lambda} illustrates the performance variations in different datasets under different settings. $\lambda$ decides how the model learns to weight the intra-class and inter-class information simultaneously. We observe that too much weight on the intra-class similarity will degrade the inter-class similarity predictions. Experiments suggest 0.05 can work well.  

\section{Conclusion}
In this work, we propose an outlier-sensitive radiography image retrieval system \textbf{OSCARS}, which goes beyond retrieving images with the most inter-class similarity but also inspects the intra-class similarity implicitly when query images show certain variations. Utilizing the automatic learning of clean internal distribution, the intra-class variations of external sources are captured and used to generate intra-class labels by splitting the class into several groups. Feeding the sampled quadruplets consisting of both the intra, inter-class positive and negative samples to the image feature learner, a weighted margin loss is adopted to optimize the retrieval network. The resulting retrieval system is sensitive to outlier-related queries as it has learnt to rank the retrieved results based on both intra-class and inter-class similarities. This outlier-sensitive image retrieval approach provides clinical users the access to receive more relevant medical images and allow radiologists to process and analyze radiography images more effectively.

\clearpage

\appendix
\onecolumn

\section{Dataset Details}
We here introduce the details about the Stanford MURA and CheXpert dataset, and the samples we used in training and evaluation. 

\textit{Stanford MURA} contains 21,471 images. We sample a quadruplet for each image and split the quadruplets into training and validation with the ratio of 8:2. We evaluate the retrieval performance with the left unseen 1873 images.  

\textit{CheXpert} in total has 223,414 training images, of which 138,358 are in frontal view. By filtering out the invalid samples, 118,286 are left. For each image, we sample a quadruplet, resulting 118,286 training quadruplets. We thus split the quadruplets into training(80\%) and validation(20\%) parts. The left 282 frontal chest X-rays are used for testing.    

Since both the two datasets are in varied sizes, we resize all the images into a fixed size of $224\times224\times3$ to fit the feature extractor network. 

\section{Sensitivity Results}
We report the sensitivity results on both MURA and CheXpert datasets. We only calculate the sensitivity for the correct hits. Therefore, even in some situations, a model has lower sensitivity values, the general evaluation of the model performance should take the recall and precision into consideration. Because the anomaly score ranges of MURA can vary a lot, we scale its score into the range of [0,1] with a sigmoid function. For CheXpert dataset, we keep the original anomaly scores in use. As CheXpert data is with multi-labels, one sample with more than one label can have multiple anomaly scores considering each class variations. Therefore, when there are multiple hits, we take the minimum difference of the anomaly scores between the query image and the database images. Compared to MURA classes, chest X-rays are often similar with each other and thus difficult to retrieve. We here present the results for CheXpert with higher float precision. Generally, the lower the sensitivity values the better. 

\begin{table*}[htp]
\caption{Sensitivity Similarity of Stanford MURA and CheXpert datasets. Best values are in bold.~\label{sensitivity_mura_sigmoid}}
\resizebox{\linewidth}{!}{%
\begin{tabular}{|l|ccccc|ccccc|}
\hline
\multirow{2}{*} {Method}& \multicolumn{5}{c|}{MURA}  & \multicolumn{5}{c|}{CheXpert}\\ \cline{2-11} 
& S@1$\downarrow$ & S@5$\downarrow$ & S@10$\downarrow$ & S@50$\downarrow$ & S@100$\downarrow$ & S@1$\downarrow$ & S@5$\downarrow$ & S@10$\downarrow$ & S@50$\downarrow$ & S@100$\downarrow$\\
\hline 
DeepRank & 0.034  & 0.035 & 0.036  & 0.039  & 0.041 & 0.01157 & 0.01252 & 0.01217  & 0.01221  & 0.01217\\ 
FastAP & 0.036 & 0.037 & 0.038  & 0.041  & 0.043 & 0.01369  & 0.01287 & 0.01281  & 0.01293  & 0.01285\\ 
MultiSimilarity & 0.035 & 0.037 & 0.038 & 0.041  & 0.043 & 0.01115  & 0.01250 & 0.01265  & 0.01303  & 0.01338\\ 
CircleLoss & 0.036 & 0.038 & 0.039 & 0.041 & 0.043 & 0.01600  & 0.01498 & 0.01529  & 0.01469  & 0.01518\\ 
SupCon & 0.038  & 0.039  & 0.040 & 0.043 & 0.045 & 0.01639  & 0.01366 & 0.01309  & 0.01348  & 0.01348\\ \hline
OSCARS \textit{(ours)} & \textbf{0.030} & \textbf{0.032}  & \textbf{0.033}  & \textbf{0.036}  & \textbf{0.038} & \textbf{0.01090}  & \textbf{0.01015} & \textbf{0.00998}  & \textbf{0.01039}  & \textbf{0.01037}\\ 
\hline 
\end{tabular}%
}
\end{table*}

\section{Visualization of MURA Retrieval}
\begin{figure*}[htp]
\centering
\includegraphics[width=\linewidth]{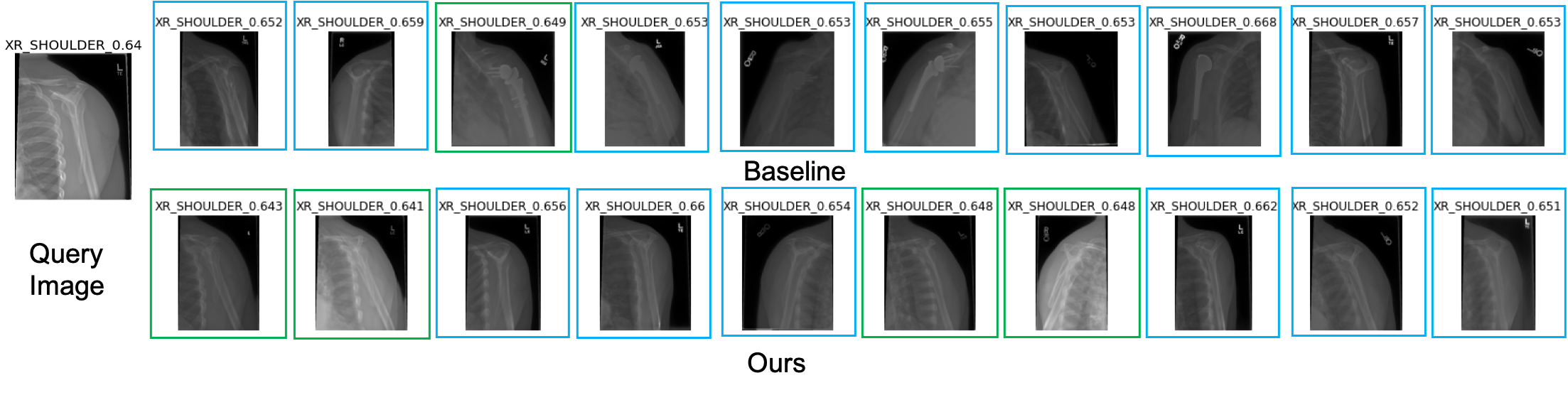}
\caption{Query results, left is the query image, retrieval results are shown in the right part. Green boxes mean both intra- and inter-class correct; blue boxes are for inter-class correct predictions. Each retrieval image has its corresponding label on top of itself, and for the correct predictions, we also put the anomaly scores on them. Closer anomaly scores mean more similarity.}
\label{mura_res1}
\end{figure*}
\begin{figure*}[h]
\centering
\includegraphics[width=\linewidth]{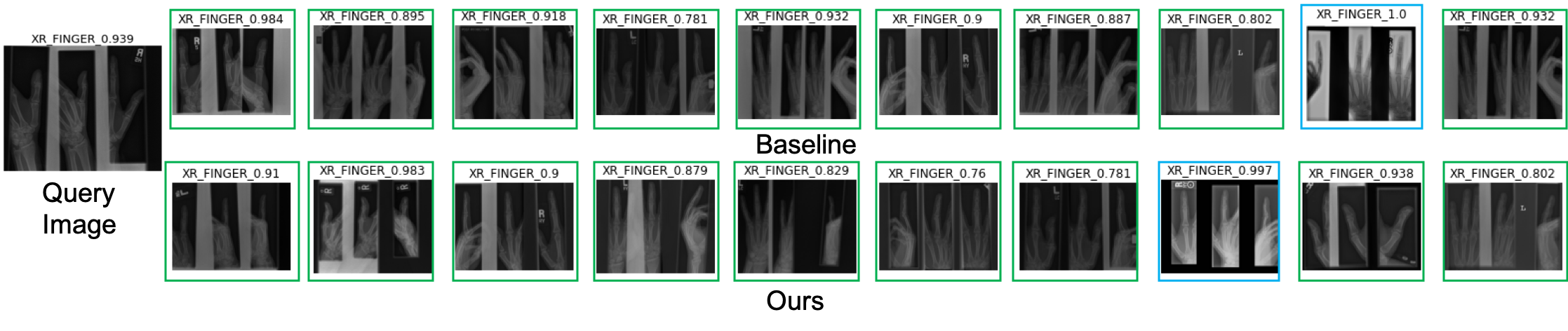}
\includegraphics[width=\linewidth]{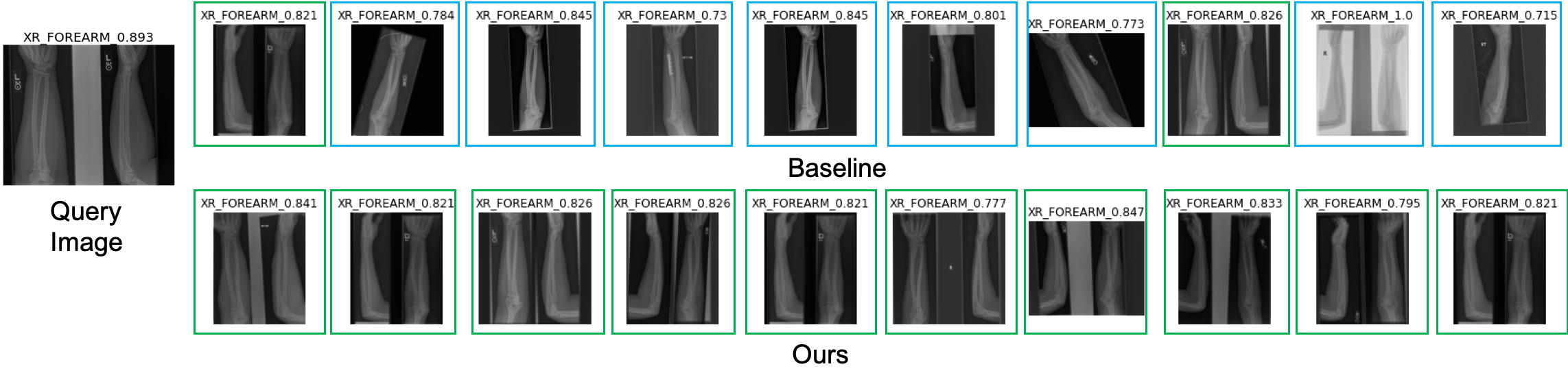}
\includegraphics[width=\linewidth]{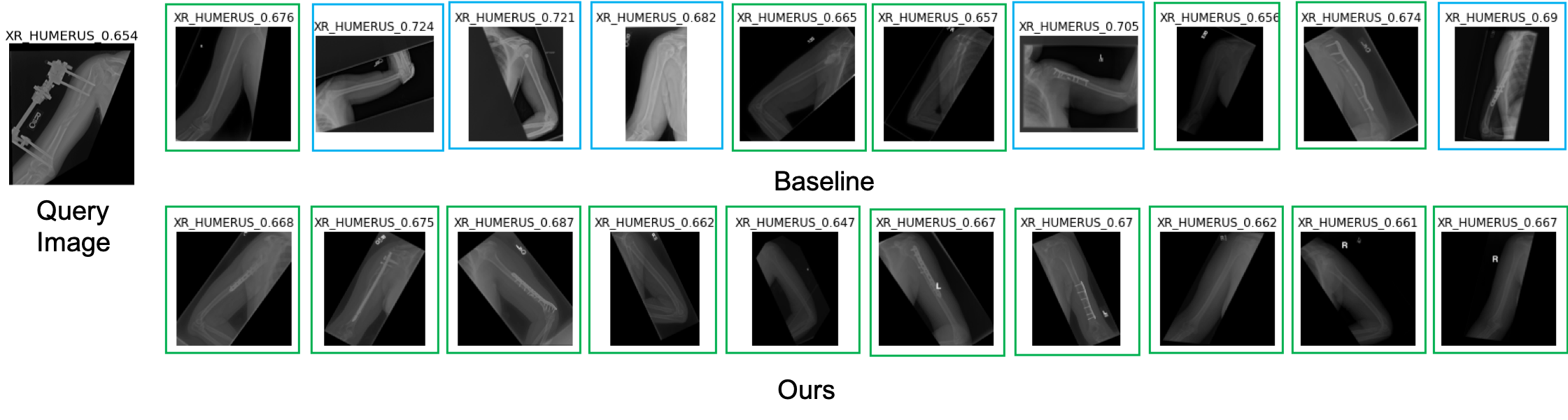}
\caption{More query results for MURA dataset. Captions follow the same style as Fig.~\ref{mura_res1}.}
\label{mura_res2}
\end{figure*}
\clearpage
\section{Visualization of CheXpert Retrieval}
Since a sample of CheXpert can have more than one lables, we encode the labels into a binary code with a length of 14, of which 1 means the data belongs to the class and 0 means irrelevant. The 14-bit label corresponds to the classes \textit{No Finding}, \textit{Enlarged Cardiomediastinum}, \textit{Cardiomegaly}, \textit{Lung Lesion}, \textit{Lung Opacity}, \textit{Edema}, \textit{Consolidation}, \textit{Pneumonia}, \textit{Atelectasis}, \textit{Pneumothorax}, \textit{Pleural Effusion}, \textit{Pleural Other}, \textit{Fracture}, \textit{Support Devices} in order. For simplicity, we only show labels, not with the anomaly scores. 
\begin{figure*}[htp]
\centering
\includegraphics[width=\linewidth]{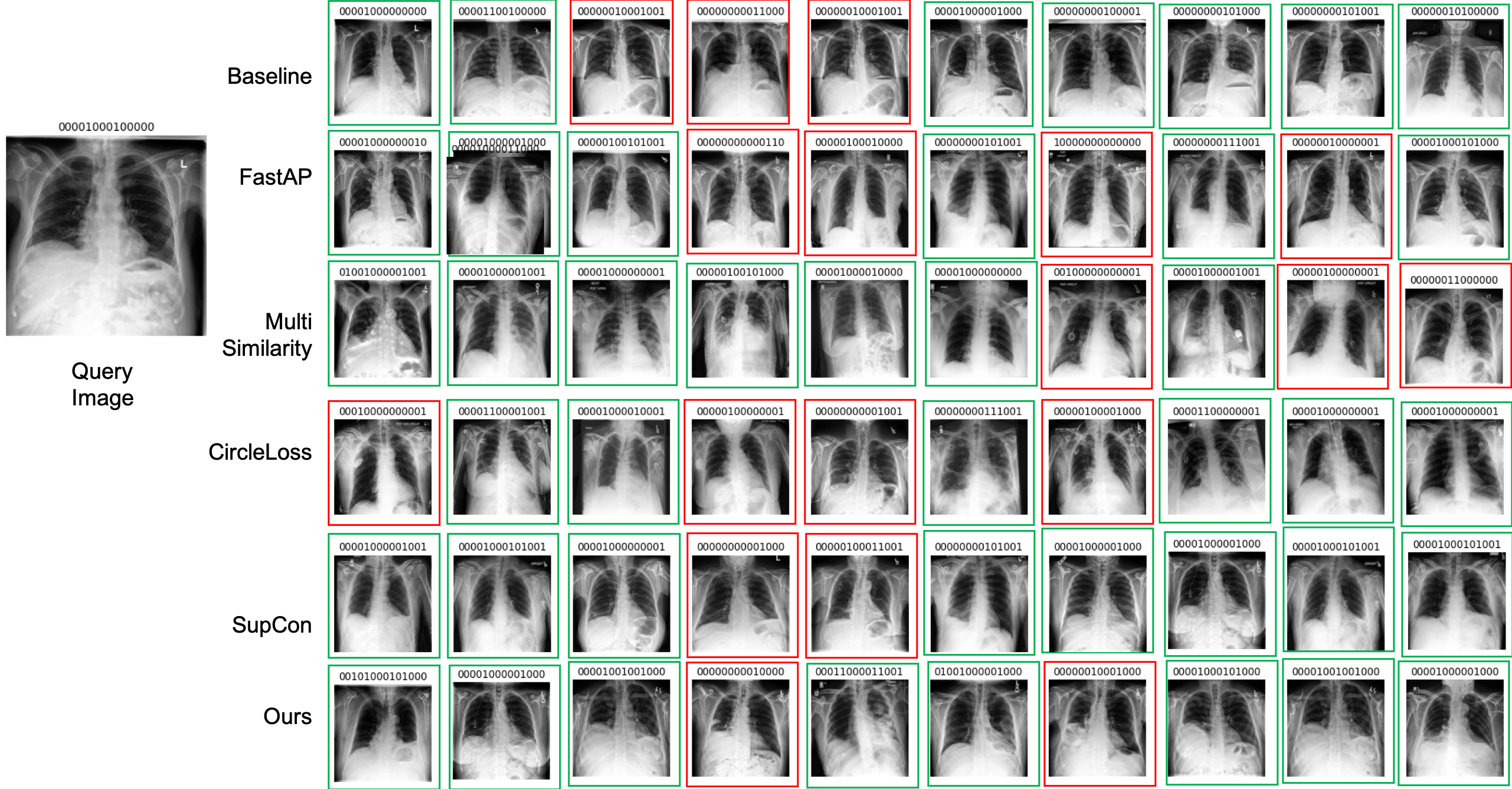}
\caption{CheXpert query results, left is the query image, retrieval results are shown in the right part. Green boxes mean both intra-class and inter-class correct; blue boxes are for inter-class correct predictions and red boxes are for wrong predictions. Each retrieval image has its corresponding label on top of itself.}
\label{mura_res}
\end{figure*}

\twocolumn




\bibliography{mybibfile}

\begin{thebibliography}{26}
\expandafter\ifx\csname natexlab\endcsname\relax\def\natexlab#1{#1}\fi
\providecommand{\url}[1]{\texttt{#1}}
\providecommand{\href}[2]{#2}
\providecommand{\path}[1]{#1}
\providecommand{\DOIprefix}{doi:}
\providecommand{\ArXivprefix}{arXiv:}
\providecommand{\URLprefix}{URL: }
\providecommand{\Pubmedprefix}{pmid:}
\providecommand{\doi}[1]{\href{http://dx.doi.org/#1}{\path{#1}}}
\providecommand{\Pubmed}[1]{\href{pmid:#1}{\path{#1}}}
\providecommand{\bibinfo}[2]{#2}
\ifx\xfnm\relax \def\xfnm[#1]{\unskip,\space#1}\fi
\bibitem[{Akg{\"u}l et~al.(2011)Akg{\"u}l, Rubin, Napel, Beaulieu, Greenspan
  and Acar}]{akgul2011content}
\bibinfo{author}{Akg{\"u}l, C.B.}, \bibinfo{author}{Rubin, D.L.},
  \bibinfo{author}{Napel, S.}, \bibinfo{author}{Beaulieu, C.F.},
  \bibinfo{author}{Greenspan, H.}, \bibinfo{author}{Acar, B.},
  \bibinfo{year}{2011}.
\newblock \bibinfo{title}{Content-based image retrieval in radiology: current
  status and future directions}.
\newblock \bibinfo{journal}{Journal of digital imaging} \bibinfo{volume}{24},
  \bibinfo{pages}{208--222}.
\bibitem[{Anavi et~al.(2015)Anavi, Kogan, Gelbart, Geva and
  Greenspan}]{anavi2015comparative}
\bibinfo{author}{Anavi, Y.}, \bibinfo{author}{Kogan, I.},
  \bibinfo{author}{Gelbart, E.}, \bibinfo{author}{Geva, O.},
  \bibinfo{author}{Greenspan, H.}, \bibinfo{year}{2015}.
\newblock \bibinfo{title}{A comparative study for chest radiograph image
  retrieval using binary texture and deep learning classification}, in:
  \bibinfo{booktitle}{2015 37th annual international conference of the IEEE
  engineering in medicine and biology society (EMBC)},
  \bibinfo{organization}{IEEE}. pp. \bibinfo{pages}{2940--2943}.
\bibitem[{Cakir et~al.(2019)Cakir, He, Xia, Kulis and Sclaroff}]{cakir2019deep}
\bibinfo{author}{Cakir, F.}, \bibinfo{author}{He, K.}, \bibinfo{author}{Xia,
  X.}, \bibinfo{author}{Kulis, B.}, \bibinfo{author}{Sclaroff, S.},
  \bibinfo{year}{2019}.
\newblock \bibinfo{title}{Deep metric learning to rank}, in:
  \bibinfo{booktitle}{Proceedings of the IEEE/CVF Conference on Computer Vision
  and Pattern Recognition}, pp. \bibinfo{pages}{1861--1870}.
\bibitem[{Chen et~al.(2022)Chen, Liu, Wang, Bakker, Georgiou, Fieguth, Liu and
  Lew}]{chen2022deep}
\bibinfo{author}{Chen, W.}, \bibinfo{author}{Liu, Y.}, \bibinfo{author}{Wang,
  W.}, \bibinfo{author}{Bakker, E.}, \bibinfo{author}{Georgiou, T.},
  \bibinfo{author}{Fieguth, P.}, \bibinfo{author}{Liu, L.},
  \bibinfo{author}{Lew, M.S.}, \bibinfo{year}{2022}.
\newblock \bibinfo{title}{Deep learning for instance retrieval: A survey}.
\newblock \href{http://arxiv.org/abs/2101.11282}{\tt arXiv:2101.11282}.
\bibitem[{Chowdhury et~al.(2016)Chowdhury, Bulo, Moreno, Kundu and
  Smedby}]{chowdhury2016efficient}
\bibinfo{author}{Chowdhury, M.}, \bibinfo{author}{Bulo, S.R.},
  \bibinfo{author}{Moreno, R.}, \bibinfo{author}{Kundu, M.K.},
  \bibinfo{author}{Smedby, {\"O}.}, \bibinfo{year}{2016}.
\newblock \bibinfo{title}{An efficient radiographic image retrieval system
  using convolutional neural network}, in: \bibinfo{booktitle}{2016 23rd
  International Conference on Pattern Recognition (ICPR)},
  \bibinfo{organization}{IEEE}. pp. \bibinfo{pages}{3134--3139}.
\bibitem[{Deng et~al.(2009)Deng, Dong, Socher, Li, Li and
  Fei-Fei}]{deng2009imagenet}
\bibinfo{author}{Deng, J.}, \bibinfo{author}{Dong, W.},
  \bibinfo{author}{Socher, R.}, \bibinfo{author}{Li, L.J.},
  \bibinfo{author}{Li, K.}, \bibinfo{author}{Fei-Fei, L.},
  \bibinfo{year}{2009}.
\newblock \bibinfo{title}{Imagenet: A large-scale hierarchical image database},
  in: \bibinfo{booktitle}{2009 IEEE conference on computer vision and pattern
  recognition}, \bibinfo{organization}{Ieee}. pp. \bibinfo{pages}{248--255}.
\bibitem[{Duan and Kuo(2021)}]{duan2021bridging}
\bibinfo{author}{Duan, J.}, \bibinfo{author}{Kuo, C.C.J.},
  \bibinfo{year}{2021}.
\newblock \bibinfo{title}{Bridging gap between image pixels and semantics via
  supervision: A survey}.
\newblock \bibinfo{journal}{arXiv preprint arXiv:2107.13757} .
\bibitem[{Dubey(2021)}]{dubey2021decade}
\bibinfo{author}{Dubey, S.R.}, \bibinfo{year}{2021}.
\newblock \bibinfo{title}{A decade survey of content based image retrieval
  using deep learning}.
\newblock \bibinfo{journal}{IEEE Transactions on Circuits and Systems for Video
  Technology} .
\bibitem[{Guo et~al.(2021a)Guo, Gichoya, Purkayastha and
  Banerjee}]{guo2021cvad}
\bibinfo{author}{Guo, X.}, \bibinfo{author}{Gichoya, J.W.},
  \bibinfo{author}{Purkayastha, S.}, \bibinfo{author}{Banerjee, I.},
  \bibinfo{year}{2021}a.
\newblock \bibinfo{title}{Cvad: A generic medical anomaly detector based on
  cascade vae}.
\newblock \bibinfo{journal}{arXiv preprint arXiv:2110.15811} .
\bibitem[{Guo et~al.(2021b)Guo, Gichoya, Trivedi, Purkayastha and
  Banerjee}]{guo2021medshift}
\bibinfo{author}{Guo, X.}, \bibinfo{author}{Gichoya, J.W.},
  \bibinfo{author}{Trivedi, H.}, \bibinfo{author}{Purkayastha, S.},
  \bibinfo{author}{Banerjee, I.}, \bibinfo{year}{2021}b.
\newblock \bibinfo{title}{Medshift: identifying shift data for medical dataset
  curation}.
\newblock \bibinfo{journal}{arXiv preprint arXiv:2112.13885} .
\bibitem[{He et~al.(2016)He, Zhang, Ren and Sun}]{he2016deep}
\bibinfo{author}{He, K.}, \bibinfo{author}{Zhang, X.}, \bibinfo{author}{Ren,
  S.}, \bibinfo{author}{Sun, J.}, \bibinfo{year}{2016}.
\newblock \bibinfo{title}{Deep residual learning for image recognition}, in:
  \bibinfo{booktitle}{Proceedings of the IEEE conference on computer vision and
  pattern recognition}, pp. \bibinfo{pages}{770--778}.
\bibitem[{Hwang et~al.(2012)Hwang, Lee and Choi}]{hwang2012medical}
\bibinfo{author}{Hwang, K.H.}, \bibinfo{author}{Lee, H.},
  \bibinfo{author}{Choi, D.}, \bibinfo{year}{2012}.
\newblock \bibinfo{title}{Medical image retrieval: past and present}.
\newblock \bibinfo{journal}{Healthcare informatics research}
  \bibinfo{volume}{18}, \bibinfo{pages}{3--9}.
\bibitem[{Irvin et~al.(2019)Irvin, Rajpurkar, Ko, Yu, Ciurea-Ilcus, Chute,
  Marklund, Haghgoo, Ball, Shpanskaya et~al.}]{irvin2019chexpert}
\bibinfo{author}{Irvin, J.}, \bibinfo{author}{Rajpurkar, P.},
  \bibinfo{author}{Ko, M.}, \bibinfo{author}{Yu, Y.},
  \bibinfo{author}{Ciurea-Ilcus, S.}, \bibinfo{author}{Chute, C.},
  \bibinfo{author}{Marklund, H.}, \bibinfo{author}{Haghgoo, B.},
  \bibinfo{author}{Ball, R.}, \bibinfo{author}{Shpanskaya, K.}, et~al.,
  \bibinfo{year}{2019}.
\newblock \bibinfo{title}{Chexpert: A large chest radiograph dataset with
  uncertainty labels and expert comparison}, in:
  \bibinfo{booktitle}{Proceedings of the AAAI conference on artificial
  intelligence}, pp. \bibinfo{pages}{590--597}.
\bibitem[{Khosla et~al.(2020)Khosla, Teterwak, Wang, Sarna, Tian, Isola,
  Maschinot, Liu and Krishnan}]{khosla2020supervised}
\bibinfo{author}{Khosla, P.}, \bibinfo{author}{Teterwak, P.},
  \bibinfo{author}{Wang, C.}, \bibinfo{author}{Sarna, A.},
  \bibinfo{author}{Tian, Y.}, \bibinfo{author}{Isola, P.},
  \bibinfo{author}{Maschinot, A.}, \bibinfo{author}{Liu, C.},
  \bibinfo{author}{Krishnan, D.}, \bibinfo{year}{2020}.
\newblock \bibinfo{title}{Supervised contrastive learning}.
\newblock \bibinfo{journal}{Advances in Neural Information Processing Systems}
  \bibinfo{volume}{33}.
\bibitem[{Layode and Rahman(2020)}]{layode2020chest}
\bibinfo{author}{Layode, O.}, \bibinfo{author}{Rahman, M.},
  \bibinfo{year}{2020}.
\newblock \bibinfo{title}{A chest x-ray image retrieval system for covid-19
  detection using deep transfer learning and denoising auto encoder}, in:
  \bibinfo{booktitle}{2020 International Conference on Computational Science
  and Computational Intelligence (CSCI)}, \bibinfo{organization}{IEEE}. pp.
  \bibinfo{pages}{1635--1640}.
\bibitem[{Lloyd(1982)}]{lloyd1982least}
\bibinfo{author}{Lloyd, S.}, \bibinfo{year}{1982}.
\newblock \bibinfo{title}{Least squares quantization in pcm}.
\newblock \bibinfo{journal}{IEEE transactions on information theory}
  \bibinfo{volume}{28}, \bibinfo{pages}{129--137}.
\bibitem[{MacQueen et~al.(1967)}]{macqueen1967some}
\bibinfo{author}{MacQueen, J.}, et~al., \bibinfo{year}{1967}.
\newblock \bibinfo{title}{Some methods for classification and analysis of
  multivariate observations}, in: \bibinfo{booktitle}{Proceedings of the fifth
  Berkeley symposium on mathematical statistics and probability},
  \bibinfo{organization}{Oakland, CA, USA}. pp. \bibinfo{pages}{281--297}.
\bibitem[{Qayyum et~al.(2017)Qayyum, Anwar, Awais and
  Majid}]{qayyum2017medical}
\bibinfo{author}{Qayyum, A.}, \bibinfo{author}{Anwar, S.M.},
  \bibinfo{author}{Awais, M.}, \bibinfo{author}{Majid, M.},
  \bibinfo{year}{2017}.
\newblock \bibinfo{title}{Medical image retrieval using deep convolutional
  neural network}.
\newblock \bibinfo{journal}{Neurocomputing} \bibinfo{volume}{266},
  \bibinfo{pages}{8--20}.
\bibitem[{Rajpurkar et~al.(2017)Rajpurkar, Irvin, Bagul, Ding, Duan, Mehta,
  Yang, Zhu, Laird, Ball et~al.}]{rajpurkar2017mura}
\bibinfo{author}{Rajpurkar, P.}, \bibinfo{author}{Irvin, J.},
  \bibinfo{author}{Bagul, A.}, \bibinfo{author}{Ding, D.},
  \bibinfo{author}{Duan, T.}, \bibinfo{author}{Mehta, H.},
  \bibinfo{author}{Yang, B.}, \bibinfo{author}{Zhu, K.},
  \bibinfo{author}{Laird, D.}, \bibinfo{author}{Ball, R.L.}, et~al.,
  \bibinfo{year}{2017}.
\newblock \bibinfo{title}{Mura: Large dataset for abnormality detection in
  musculoskeletal radiographs}.
\newblock \bibinfo{journal}{arXiv preprint arXiv:1712.06957} .
\bibitem[{Revaud et~al.(2019)Revaud, Almaz{\'a}n, Rezende and
  Souza}]{revaud2019learning}
\bibinfo{author}{Revaud, J.}, \bibinfo{author}{Almaz{\'a}n, J.},
  \bibinfo{author}{Rezende, R.S.}, \bibinfo{author}{Souza, C.R.d.},
  \bibinfo{year}{2019}.
\newblock \bibinfo{title}{Learning with average precision: Training image
  retrieval with a listwise loss}, in: \bibinfo{booktitle}{Proceedings of the
  IEEE/CVF International Conference on Computer Vision}, pp.
  \bibinfo{pages}{5107--5116}.
\bibitem[{Sotomayor et~al.(2021)Sotomayor, Mendoza, Casta{\~n}eda, Far{\'\i}as,
  Molina, Pereira, H{\"a}rtel, Solar and Araya}]{sotomayor2021content}
\bibinfo{author}{Sotomayor, C.G.}, \bibinfo{author}{Mendoza, M.},
  \bibinfo{author}{Casta{\~n}eda, V.}, \bibinfo{author}{Far{\'\i}as, H.},
  \bibinfo{author}{Molina, G.}, \bibinfo{author}{Pereira, G.},
  \bibinfo{author}{H{\"a}rtel, S.}, \bibinfo{author}{Solar, M.},
  \bibinfo{author}{Araya, M.}, \bibinfo{year}{2021}.
\newblock \bibinfo{title}{Content-based medical image retrieval and intelligent
  interactive visual browser for medical education, research and care}.
\newblock \bibinfo{journal}{Diagnostics} \bibinfo{volume}{11},
  \bibinfo{pages}{1470}.
\bibitem[{Sun et~al.(2020)Sun, Cheng, Zhang, Zhang, Zheng, Wang and
  Wei}]{sun2020circle}
\bibinfo{author}{Sun, Y.}, \bibinfo{author}{Cheng, C.}, \bibinfo{author}{Zhang,
  Y.}, \bibinfo{author}{Zhang, C.}, \bibinfo{author}{Zheng, L.},
  \bibinfo{author}{Wang, Z.}, \bibinfo{author}{Wei, Y.}, \bibinfo{year}{2020}.
\newblock \bibinfo{title}{Circle loss: A unified perspective of pair similarity
  optimization}, in: \bibinfo{booktitle}{Proceedings of the IEEE/CVF Conference
  on Computer Vision and Pattern Recognition}, pp. \bibinfo{pages}{6398--6407}.
\bibitem[{Thorndike(1953)}]{thorndike1953belongs}
\bibinfo{author}{Thorndike, R.L.}, \bibinfo{year}{1953}.
\newblock \bibinfo{title}{Who belongs in the family?}
\newblock \bibinfo{journal}{Psychometrika} \bibinfo{volume}{18},
  \bibinfo{pages}{267--276}.
\bibitem[{Wang et~al.(2014)Wang, Song, Leung, Rosenberg, Wang, Philbin, Chen
  and Wu}]{wang2014learning}
\bibinfo{author}{Wang, J.}, \bibinfo{author}{Song, Y.}, \bibinfo{author}{Leung,
  T.}, \bibinfo{author}{Rosenberg, C.}, \bibinfo{author}{Wang, J.},
  \bibinfo{author}{Philbin, J.}, \bibinfo{author}{Chen, B.},
  \bibinfo{author}{Wu, Y.}, \bibinfo{year}{2014}.
\newblock \bibinfo{title}{Learning fine-grained image similarity with deep
  ranking}, in: \bibinfo{booktitle}{Proceedings of the IEEE conference on
  computer vision and pattern recognition}, pp. \bibinfo{pages}{1386--1393}.
\bibitem[{Wang et~al.(2019)Wang, Han, Huang, Dong and Scott}]{wang2019multi}
\bibinfo{author}{Wang, X.}, \bibinfo{author}{Han, X.}, \bibinfo{author}{Huang,
  W.}, \bibinfo{author}{Dong, D.}, \bibinfo{author}{Scott, M.R.},
  \bibinfo{year}{2019}.
\newblock \bibinfo{title}{Multi-similarity loss with general pair weighting for
  deep metric learning}, in: \bibinfo{booktitle}{Proceedings of the IEEE/CVF
  Conference on Computer Vision and Pattern Recognition}, pp.
  \bibinfo{pages}{5022--5030}.
\bibitem[{Zhong et~al.(2021)Zhong, Li, Wu, Ren, Kim, Kim, Buch, Neumark, Bizzo,
  Tak et~al.}]{zhong2021deep}
\bibinfo{author}{Zhong, A.}, \bibinfo{author}{Li, X.}, \bibinfo{author}{Wu,
  D.}, \bibinfo{author}{Ren, H.}, \bibinfo{author}{Kim, K.},
  \bibinfo{author}{Kim, Y.}, \bibinfo{author}{Buch, V.},
  \bibinfo{author}{Neumark, N.}, \bibinfo{author}{Bizzo, B.},
  \bibinfo{author}{Tak, W.Y.}, et~al., \bibinfo{year}{2021}.
\newblock \bibinfo{title}{Deep metric learning-based image retrieval system for
  chest radiograph and its clinical applications in covid-19}.
\newblock \bibinfo{journal}{Medical Image Analysis} \bibinfo{volume}{70},
  \bibinfo{pages}{101993}.

\end{thebibliography}

\end{document}